\documentclass[final]{cvpr}

\usepackage{times}
\usepackage{epsfig}
\usepackage{graphicx}
\usepackage{amsmath}
\usepackage{amssymb}
\usepackage{microtype}      
\usepackage{mathrsfs}  
\usepackage{bm}
\usepackage{amssymb,amsmath}
\usepackage{multicol,multicap,graphicx}
\usepackage{graphicx}
\usepackage{subcaption}
\usepackage{wrapfig}
\usepackage{picinpar}
\usepackage{cutwin}
\usepackage{stmaryrd}
\usepackage{graphicx,multicol}
\usepackage{algorithm}  
\usepackage{algorithmic}
\usepackage{amsfonts}
\usepackage{amsfonts,amssymb} 
\usepackage{verbatim}
\usepackage{url}

\DeclareMathOperator*{\argmin}{arg\,min}
\DeclareMathOperator*{\argmax}{arg\,max}
\usepackage{booktabs}
\usepackage{multirow}

\usepackage[pagebackref=true,breaklinks=true,colorlinks,bookmarks=false]{hyperref}

\def\cvprPaperID{63} 
\def\confYear{L2ID 2021}

\begin{document}

\title{Efficient Pre-trained Features and Recurrent Pseudo-Labeling 
\\in Unsupervised Domain Adaptation}

\author{Youshan Zhang \ \ \ \ \ \ \ \  Brian D.\ Davison \\
Computer Science and Engineering, Lehigh University, Bethlehem, PA, USA\\
{ \{yoz217, bdd3\}@lehigh.edu}
}

\maketitle
\thispagestyle{empty}
\pagestyle{empty}
\begin{abstract}
Domain adaptation (DA) mitigates the domain shift problem when transferring knowledge from one annotated domain to another similar but different unlabeled domain. However, existing models often utilize one of the ImageNet models as the backbone without exploring others, and fine-tuning or retraining the backbone ImageNet model is also time-consuming. Moreover, pseudo-labeling has been used to improve the performance in the target domain, while how to generate confident pseudo labels and explicitly align domain distributions has not been well addressed.  In this paper, we show how to efficiently opt for the best pre-trained features from seventeen well-known ImageNet models in unsupervised DA problems. In addition, we propose a recurrent pseudo-labeling model using the best pre-trained features (termed PRPL) to improve classification performance. To show the effectiveness of PRPL, we evaluate it on three benchmark datasets, Office+Caltech-10, Office-31, and Office-Home. Extensive experiments show that our model reduces computation time and boosts the mean accuracy to 98.1\%, 92.4\%, and 81.2\%, respectively, substantially outperforming the state of the art.
\end{abstract}

\section{Introduction}
With the explosive growth of information in the current era, there are massive amounts of data from multiple sources and corresponding to varied scenarios. However, not all tasks have enough annotated data for training, and collecting sufficient labeled data is a big investment of time and effort. Therefore, in order to build machine learning models it is often necessary to transfer knowledge from one labeled domain to an unlabeled domain. Due to dataset bias or domain shift~\cite{pan2010survey}, the generalization ability of the learned model on the unlabeled domain has been severely compromised. Domain adaptation (DA) is proposed to circumvent the domain shift problem.

Unsupervised domain adaptation (UDA) transfers knowledge learned from a  label-rich source domain to a fully unlabeled target domain~\cite{long2016unsupervised}. Most prior methods focus on matching (marginal, conditional, and joint) distributions between two domains to learn domain-invariant representations. Maximum Mean Discrepancy (MMD) is one of the most popular distance metrics when minimizing differences between two distributions. Long et al.~\cite{long2015learning} proposed a Deep Adaptation Network (DAN) that considered multiple kernels of MMD functions. Recently, 
Kang et al.~\cite{kang2019contrastive} extended MMD to the contrastive domain discrepancy loss. 
However, these distance-based metrics can also mix samples of different classes together. 
Recently, adversarial learning has shown its power in learning domain invariant representations. The domain discriminator aims to distinguish the source domain from the target domain, while the feature extractor aims to learn domain-invariant representations to fool the domain discriminator~\cite{kang2019contrastive}. 
Sometimes, pseudo-labeling is proposed to learn the target discriminative representations~\cite{xie2018learning,zhang2018collaborative}. However, the credibility of these pseudo labels is unknown. 

To address the above challenges, this paper provides two specific contributions:

\begin{enumerate}
\vspace{-0.2cm}
    \item To reduce computation time, we extract features from seventeen pre-trained ImageNet models and then design a fast and efficient unsupervised metric to select the best pre-trained features for the domain transfer tasks. 

    \item We develop a recurrent pseudo-labeling paradigm to continuously select high confidence transfer examples from the target domain and minimize the marginal and conditional discrepancies between the two domains.
\end{enumerate}
\vspace{-0.2cm}
We conduct extensive experiments on three benchmark datasets (Office + Caltech-10, Office-31, and Office-Home),
achieving higher 
accuracy than state-of-the-art methods.

\section{Related work}
\paragraph{Pre-training.}

Pre-training is one of the dominant components of transfer learning. Recent deep networks often apply a pre-trained network (typically trained on the ImageNet dataset) as the initialization for object recognition and segmentation.  As in many computer vision tasks, it is often slow and tedious to train a new network from scratch. Hence, using a pre-trained model on one dataset to help another is 
a major advantage of 
transfer learning. Traditional DA methods relied on the extracted features from the pre-trained ImageNet models and then aligned the marginal or conditional distributions between different domains~\cite{zhang2019modified,zhang2019transductive,zhang2020domain}. Recent deep networks frequently select ResNet50 as the backbone network for UDA~\cite{wang2019transferable,long2018conditional}. Notably, 
other deep networks are not investigated,
even though
different ImageNet models affect the performance of UDA on traditional methods~\cite{zhang2020impact,zhang2021adversarial}. 
The impact of extracted pre-trained features on deep networks are not well explored. In addition, selecting the best pre-trained features based on performance in the target domain requires significant computation to train models in the supervised source domain and infer to the unlabeled target.


\paragraph{Pseudo-labeling.}
Pseudo-labeling is another technique to address UDA and also achieves substantial performance on multiple tasks. Pseudo-labeling typically generates pseudo labels for the target domain based on the predicted class probability~\cite{saito2017asymmetric,xie2018learning,zhang2018collaborative,chen2019progressive,zhang2021adversarial2}. Under such a regime, some target domain label information can be considered during training. In deep networks, the source classifier is usually treated as an initial pseudo labeler to generate the pseudo labels (and use them as if they were real labels). Different algorithms are proposed to obtain additional pseudo labels and promote distribution alignment between the two domains.

An asymmetric tri-training method for UDA has been proposed to generate pseudo labels for target samples using two networks, and the third  can learn from them to obtain target discriminative representations~\cite{saito2017asymmetric}. Xie et al.~\cite{xie2018learning} proposed a  Moving Semantic Transfer Network (MSTN) to develop semantic matching and domain adversary losses to obtain pseudo labels. Zhang et al.~\cite{zhang2018collaborative} designed a new criterion to select pseudo-labeled target samples and developed an iterative approach called incremental CAN (iCAN), in which they select samples iteratively and retrain the network using the expanded training set. Progressive Feature Alignment Network (PFAN)~\cite{chen2019progressive}  aligns the discriminative features across domains progressively and employs an easy-to-hard transfer strategy for iterative learning. Chang et al.~\cite{chang2019domain} proposed to combine the external UDA algorithm and the proposed domain-specific batch normalization to estimate the pseudo labels of samples in the target domain and more effectively learn the domain-specific features. Constrictive Adaptation Network (CAN) also employed batch normalization layers to capture the domain-specific distributions~\cite{kang2019contrastive}.

These methods highly rely on the pseudo labels to compensate for the lack of categorical information in the target domain. However, they did not check the quality of pseudo-labels, as noisy pseudo-labeled samples hurt model performance. In addition, most pseudo-labeling methods employ a two-stage paradigm. The pseudo labels in the first stage are generated and then used to train the model along with the labeled source domain. Differing from previous work~\cite{long2015learning,chen2019progressive}, we recurrently generate high confidence examples using a novel scheme.  

\section{Methodology}
\vspace{-0.1cm}
\subsection{Problem}
Here we discuss the unsupervised  domain adaptation (UDA) problem  and introduce some basic notation. Given a labeled source domain $\mathcal{D_S} = \{\mathcal{X}_\mathcal{S}^i, \mathcal{Y}^i_\mathcal{S} \}_{i=1}^\mathcal{N_S}$ with $\mathcal{N_S}$ samples in $C$ categories and an unlabeled target domain $\mathcal{D_T} = \{\mathcal{X}_\mathcal{T}^j\}_{j=1}^{\mathcal{N_T}}$ with $\mathcal{N_T}$ samples in the same $C$ categories  ($\mathcal{Y_T}$ for evaluation only), our challenge is how to get a well-trained classifier so that domain discrepancy is minimized and generalization error in the target domain is reduced.

In UDA, a typical method would first select one of the ImageNet models as the backbone network for feature extraction. Then, more fine-tuning layers and loss functions are added to minimize the discrepancy between $\mathcal{X_S}$ and $\mathcal{X_T}$. In addition, most pseudo-labeling algorithms do not check the reliability of the generated pseudo labels.  These approaches face two critical limitations: (1) they did not explore other pre-trained networks and did not select the best pre-trained features. It is hence necessary to develop an unsupervised metric to quickly determine the best pre-trained features among different ImageNet models. (2) Noisy pseudo labels can deteriorate domain invariant features, and the conditional distributions of two domains are difficult to align at the category level, and the marginal and conditional distributions are not jointly well aligned.

To mitigate these shortcomings, we propose a recurrent pseudo-labeling using the best pre-trained features (PRPL) model. We also jointly align the marginal and conditional distributions of the two domains.

\subsection{Pre-training Feature Representation}

Feature extraction is an easy and fast way to use the power of deep learning without investing resources into training or fine-tuning a network. It would be especially useful when there are no powerful GPUs to train additional deep networks. However, one disadvantage of feature extraction using pre-trained networks is that it has lower performance than fine-tuning the same network since feature extraction is a single pass over the images. However, previous work~\cite{zhang2020impact} suggests that a better ImageNet model will produce better features for UDA. Therefore, we can extract features from a better ImageNet model to compensate for the lower performance of not fine-tuning. Moreover, feature extraction is significantly faster than fine-tuning a neural network. Thus, our design goal in feature representation learning is to find fast and accurate features from  well-trained ImageNet models.

\begin{figure}[t]
\centering
\includegraphics[width=1\columnwidth]{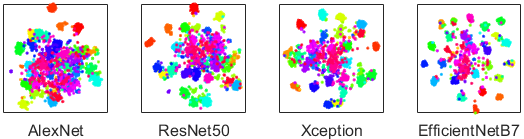}
\caption{t-SNE view of extracted features from four pre-trained networks (AlexNet~\cite{krizhevsky2012imagenet}, ResNet50~\cite{he2016deep}, Xception~\cite{chollet2017xception} and EfficientNetB7~\cite{tan2019efficientnet}). Different colors
represent different classes.  EfficientNetB7 has better features than others since the classes are more separate (from Amazon domain in Office31 dataset). }
\vspace{-0.3cm}
\label{fig:tsne_f}
\end{figure}

Since there are several well-trained ImageNet models, we employ a feature extractor $\Phi$ to extract features from the source and target images using a pre-trained model\footnote{$\Phi$ extracts features from the layer prior to last fully connected layer of our examined pre-trained models.}. Fig.~\ref{fig:tsne_f} shows extracted features using four different pre-trained ImageNet models. EfficientNetB7~\cite{tan2019efficientnet} has better performance and better-separated features than others from visual inspection. However, we should have an algorithmic solution that chooses the best pre-trained features. We thus design a fast and accurate unsupervised metric. For a given feature extractor $\Phi_k$, the mean distance between latent represented source $\Phi_k(\mathcal{X}_\mathcal{S})$ and target domain $\Phi_k(\mathcal{X}_\mathcal{T})$ can be denoted as: 
\begin{equation}\label{eq:pre}
  Dist_k^{Pre}= || \frac{1}{\mathcal{N}_\mathcal{S}} \sum_{i=1}^{\mathcal{N}_\mathcal{S}} \Phi_k(\mathcal{X}_\mathcal{S}^i) - \frac{1}{\mathcal{N}_\mathcal{T}} \sum_{j=1}^{\mathcal{N}_\mathcal{T}} \Phi_k(\mathcal{X}_\mathcal{T}^j) ||_2.
\end{equation}
where $\Phi_k$ is the $k^{th} \in \{1, 2, \cdots K\}$ feature extractor from seventeen pre-trained models ($K=17$) and $||\cdot||_2$ is the L2 norm. With a different $\Phi_k$, such a distance can be varied. $Dist_k^{Pre}$ is an easy and fast unsupervised metric to quantify the quality of extracted pre-trained features. 

Therefore, we can select the best pre-trained features if $Dist_k^{Pre}$ has the shortest distance between two domains. The performance of different pre-trained feature distances is shown in Sec~\ref{sec:pre}. The $\Phi$ in the following section refers to the best feature extractor, \textit{i.e.}, EfficientNetB7. 

\subsection{Feature Alignment}
\vspace{-0.2cm}
\subsubsection{Initial Source Classifier}
\begin{figure*}[h]
\centering
\includegraphics[width=1.8\columnwidth]{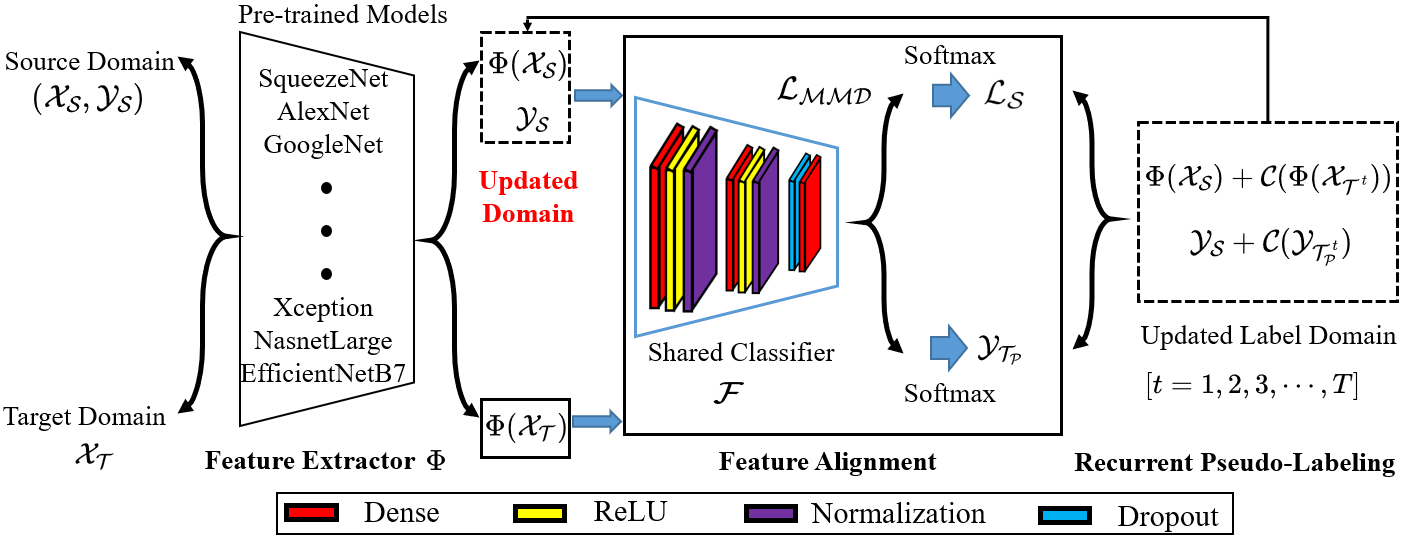}
\caption{Architecture of the PRPL model. We first extract features $\Phi(\mathcal{X_{S/T}})$ for both source and target domains via $\Phi$ using a pre-trained model, and then train the shared classifier $\mathcal{F}$. In recurrent pseudo-labeling, confident pseudo-labeled examples ($\{\mathcal{C} (\Phi(\mathcal{X}_{\mathcal{T}^{t}})), \mathcal{C} (\mathcal{Y}_{\mathcal{T}_{\mathcal{P}}^{t}})\}$) are generated continuously in each $t$ to form the updated label domain. During  training, the updated label domain will keep replacing 
$\{\Phi(\mathcal{X_S}), \mathcal{Y_S}\}$ (the rectangle above updated domain).  $\mathcal{L_S}$ is  source classification loss and $\mathcal{L_{MMD}}$ is maximum mean  discrepancy loss.  Best viewed in color.}  
\vspace{-0.3cm}
\label{fig:model}
\end{figure*}

The task in the source domain is to minimize the typical cross-entropy loss in the following equation:
\begin{equation}\label{eq:lc}
    \mathcal{L_S} = - \frac{1}{\mathcal{N}_\mathcal{S}}\sum_{i=1}^{\mathcal{N}_\mathcal{S}} \sum_{c=1}^{C} \mathcal{Y}_{\mathcal{S}_{c}}^{i} \text{log}  (\mathcal{F}_c(\Phi(\mathcal{X}_{\mathcal{S}}^{i}))), 
\end{equation}
where $\mathcal{Y}_{\mathcal{S}_{c}}^{i} \in [0, 1]^{C}$ is the binary indicator of each class $c$ in true label for observation $\Phi(\mathcal{X}_{\mathcal{S}}^{i})$, and $\mathcal{F}_c(\Phi(\mathcal{X}_{\mathcal{S}}^{i}))$ is the predicted probability of class $c$ (using the softmax function as shown in Fig.~\ref{fig:model}).

\vspace{-0.2cm}
\subsubsection{Maximum Mean Discrepancy}
Maximum mean discrepancy (MMD)~\cite{long2016unsupervised} is a non-parametric distance measure to compare the distributions of source and target domains by mapping data into reproducing kernel Hilbert space. 
After the initial classifier $\mathcal{F}$,  it is expressed by 
\begin{gather}\label{eq:mmd1}
\scalebox{0.93}{$
\begin{aligned}
     \mathcal{L_{MMD}}   & =   \frac{1}{\mathcal{N}_\mathcal{S}^2} \sum_{i, j}^{\mathcal{N_S}} \kappa(L_{\mathcal{S}}^{i}, L_{\mathcal{S}}^{j})  \ +  \frac{1}{\mathcal{N}_\mathcal{T}^2} \sum_{i, j}^{\mathcal{N_T}} \kappa(L_{\mathcal{T}}^{i}, L_{\mathcal{T}}^{j})  \\ &-  \frac{2}{\mathcal{N_S}\cdot \mathcal{N_T}} \sum_{i, j}^{\mathcal{N_S},\mathcal{N_T}} \kappa(L_{\mathcal{S}}^{i}, L_{\mathcal{T}}^{j}),
\end{aligned}$}
\end{gather}
where $\kappa$ is the mean of a linear combination of multiple RBF kernels and $L_\mathcal{S} =\mathcal{F}(\Phi(\mathcal{X_S}))$, and $L_\mathcal{T} =\mathcal{F}(\Phi(\mathcal{X_T}))$. Therefore, the $\mathcal{L_{MMD}}$ aims to minimize the marginal distance between two domains defined as follows:
\begin{equation}\label{eq:dis_m}
    Dist^{Ma}= || \frac{1}{\mathcal{N}_\mathcal{S}} \sum_{i=1}^{\mathcal{N}_\mathcal{S}} L_\mathcal{S}^i - \frac{1}{\mathcal{N}_\mathcal{T}} \sum_{j=1}^{\mathcal{N}_\mathcal{T}} L_\mathcal{T}^j ||_2.
\end{equation}
\vspace{-0.5cm}
\subsection{Recurrent Pseudo-Labeling Learning}
\vspace{-0.2cm}
Initial feature alignment only trains the target domain in an unsupervised fashion. To get reliable predicted target domain labels, we train the networks with the instances of the labeled source and pseudo labeled target domains. 
Pseudo labels of target domain examples will be treated as if they are true labels. The domain invariant features are effective and better adapted in such a paradigm.

\vspace{-0.3cm}
\subsubsection{Confident Pseudo-Labeling}
In this stage, we take advantage of the initial source classifier $\mathcal{F}$ to generate confident pseudo labels and examples for the target domain. In contrast to Hinton et al.~\cite{hinton2015distilling}, we do not use the weighted sum of the soft posteriors and the one-hot hard label to train the model since the prediction of the target domain is not accurate when the classification decision from $\mathcal{F}$ is incorrect. In recurrent pseudo-labeling learning, we continuously bring confident examples and pseudo labels from the target domain to the source domain. A confident pseudo label is defined as
\begin{equation*}
    \begin{aligned}\label{eq:y_c}
        \mathcal{C} (\mathcal{Y}_{\mathcal{T}_{\mathcal{P}}}^j)  = &  \argmax_{\substack{c \in C}} \{ \mathcal{F}_c (\Phi(\mathcal{X}_{\mathcal{T}}^{j}))\} \ \ \text{if}  \max(\mathcal{F}_c (\Phi(\mathcal{X}_{\mathcal{T}}^{j}))) > p,
    \end{aligned}
\end{equation*}
where $\mathcal{F}_c (\Phi(\mathcal{X}_{\mathcal{T}}^{j}))$  is the predicted probability of class $c$ of $\Phi(\mathcal{X}_{\mathcal{T}}^{j})$, $\max(\mathcal{F}_c (\Phi(\mathcal{X}_{\mathcal{T}}^{j})))$ is probability of the dominant class, and it should be greater than probability threshold $p$ $(0\leq p \leq 1)$ for a confident pseudo label. Also, its corresponding observation $\Phi(\mathcal{X}_{\mathcal{T}}^{j})$ is called a confident example and denoted as $ \mathcal{C} (\Phi(\mathcal{X}_{\mathcal{T}}^{j}))$, where $\mathcal{C}$ means confident pseudo labels or examples. The advantage of such confident examples and pseudo labels is to support the high quality of predicted target labels and mitigate negative transfer of $\mathcal{F}$.

Therefore, we construct an updated label domain $\mathcal{D_U} = \{\mathcal{X}_{\mathcal{U}}^n, \mathcal{Y}_{\mathcal{U}}^n \}_{n=1}^{\mathcal{N_U}}$, which consists of the labeled source domain and confident target examples with pseudo labels, where $\mathcal{N_U} \leq \mathcal{N_S}+ \mathcal{N_T}$, $\mathcal{X}_{\mathcal{U}} = \Phi(\mathcal{X_S}) +  \mathcal{C} (\Phi(\mathcal{X}_{\mathcal{T}})) $ and $ \mathcal{Y}_{\mathcal{U}} =  \mathcal{Y_S} + \mathcal{C} (\mathcal{Y}_{\mathcal{T}_{\mathcal{P}}})$, and $\mathcal{N_U}$ is controlled by $p$. $\mathcal{N_U} = 0$  if $p = 1$, and $\mathcal{N_U} = \mathcal{N_S}+ \mathcal{N_T}$ if $p = 0$.

\subsubsection{Recurrent learning}
Most existing pseudo-labeling methods only generate pseudo labels in a single iteration. However, such a paradigm cannot guarantee reliable predictions of the target domain. Therefore, we propose recurrent pseudo-labeling to continuously generate confident examples for $T$ iterations.  In each iteration $t$, the updated label domain becomes:  $\mathcal{D}_\mathcal{U}^{t} = \{\mathcal{X}_{\mathcal{U}^{t}}^n, \mathcal{Y}_{\mathcal{U}^{t}}^n \}_{n=1}^{\mathcal{N}_{\mathcal{U}^{t}}}$, where $\mathcal{X}_{\mathcal{U}^{t}} = \Phi(\mathcal{X_S}) +  \mathcal{C} (\Phi(\mathcal{X}_{\mathcal{T}^{t}})) $ and $ \mathcal{Y}_{\mathcal{U}^{t}} =  \mathcal{Y_S} + \mathcal{C} (\mathcal{Y}_{\mathcal{T}_{\mathcal{P}}^{t}})$, and $t \in \{1, 2, 3, \cdots, T\}$. To suppress potentially noisy pseudo labels, in each iteration $t$, the sample size of updated labels is always not greater than $\mathcal{N_S} + \mathcal{N_T}$, \textit{i.e.,} $\mathcal{N}_{\mathcal{U}^{t}} \leq \mathcal{N_S} + \mathcal{N_T}$. 
\begin{gather}\label{eq:y_c2}
\scalebox{0.93}{$
\begin{aligned}
        \mathcal{C} (\mathcal{Y}_{\mathcal{T}_{\mathcal{P}}^{t}})  = &  \argmax_{\substack{c \in C}} \{ \mathcal{F}_c (\Phi(\mathcal{X}_{\mathcal{T}^t}^{j}))\} \ \ \text{if}   \max(\mathcal{F}_c (\Phi(\mathcal{X}_{\mathcal{T}^t}^{j}))) > p_t,
\end{aligned}$}
\end{gather}
In addition, $\mathcal{C} (\mathcal{Y}_{\mathcal{T}_{\mathcal{P}}^{t}})$ is also updated using Eq.~\ref{eq:y_c2} for the probability threshold $p_t$, and should maintain the condition of $p_{t+1} \geq p_t (0\leq p_t\leq 1 )$ since we only want to generate reliable  pseudo labels, which will further avoid negative transfer via pushing the decision boundary toward to the target domain. In all iterations, we have a sequence of $p_t$, and $p_T=\{ p_t\}_{t=1}^T$. Therefore, we can produce confident examples and pseudo labels in each recurrent training. 

During training, the updated label domain $\mathcal{D}_\mathcal{U}^{t}$ will keep replacing the original latent represented source domain  $\{\Phi(\mathcal{X_S}), \mathcal{Y_S}\}$. The parameters in the feature alignment module will be updated via both labeled source domain and pseudo labeled target domain. Therefore, the loss function in each training iteration is given by:
\begin{equation}
    \mathcal{L}_{\mathcal{U}}^{t} =  \mathcal{L}_{\mathcal{S}}^{t} + \mathcal{L}_{\mathcal{MMD}}^{t},
\end{equation}
where
$
    \mathcal{L}_{\mathcal{S}}^{t} = - \frac{1}{\mathcal{N}_{\mathcal{U}^t}}\sum_{n=1}^{\mathcal{N}_{\mathcal{U}^t}} \sum_{c=1}^{C} \mathcal{Y}_{\mathcal{U}_{c}^t}^{n} \text{log}  (\mathcal{F}_c(\mathcal{X}_{\mathcal{U}^t}^{n})), 
$
\vspace{-0.3cm}
\begin{gather*}\label{eq:mmd2}
\scalebox{0.93}{$
\begin{aligned}
    \mathcal{L}_{\mathcal{MMD}}^{t}   & =   \frac{1}{\mathcal{N}_{\mathcal{U}^t}^2} \sum_{n, j}^{\mathcal{N}_{\mathcal{U}^t}} \kappa(L_{\mathcal{U}^t}^{n}, L_{\mathcal{U}^t}^{j})  \ +  \frac{1}{\mathcal{N}_\mathcal{T}^2}   \sum_{n, j}^{\mathcal{N_T}} \kappa(L_{\mathcal{T}}^{n}, L_{\mathcal{T}}^{j}) \\& - \frac{2}{\mathcal{N}_{\mathcal{U}^t}\cdot \mathcal{N_T}} \sum_{n, j}^{\mathcal{N}_{\mathcal{U}^t},\mathcal{N_T}} \kappa(L_{\mathcal{U}^t }^{n}, L_{\mathcal{T}}^{j}).
\end{aligned}$}
\end{gather*}
In $\mathcal{L}_{\mathcal{S}}^{t}$, $\mathcal{Y}_{\mathcal{U}_{c}^t}^{n} \in [0, 1]^{C}$ is the binary indicator of each class $c$  for observation $\mathcal{X}_{\mathcal{U}^t}^{n}$ in the $t^{th}$ training iteration, and $\mathcal{F}_c(\mathcal{X}_{\mathcal{U}^t}^{n})$ is the predicted probability of class $c$.  In $\mathcal{L}_{\mathcal{MMD}}^{t}$, $L_{\mathcal{U}^t} =\mathcal{F}(\mathcal{X}_{\mathcal{U}^t})$, $L_\mathcal{T} =\mathcal{F}(\Phi(\mathcal{X_T}))$,  and it also measures the discrepancy between updated label domain and the target domain in each $t$. Unlike Eq.~\ref{eq:mmd1}, $\mathcal{L}_{\mathcal{MMD}}^{t}$ also includes the confident examples from the target domain. The networks will be jointly optimized using the updated domain, which is equivalent to minimizing the following conditional distance during training.
\begin{equation}\label{eq:dis_c}
\begin{aligned}
   & Dist_t^{Co}  = Dist \sum_{c=1}^{C}(\mathcal{Y_{S^\textit{c}}}\vert \mathcal{X_{S^\textit{c}}}, \mathcal{C} (\mathcal{Y}_{\mathcal{T}_{\mathcal{P}}^{tc}})\vert  \mathcal{C}(\mathcal{X}_{\mathcal{T}^{tc}})) \\ & 
    = \frac{1}{C}\sum_{c=1}^{C}  || \frac{1}{\mathcal{N}_\mathcal{S}^{c}} \sum_{i=1}^{\mathcal{N}_\mathcal{S}^c} \Phi(\mathcal{X}_\mathcal{S^\textit{c}}^i)- \frac{1}{\mathcal{C} (\mathcal{N}_{\mathcal{T}^t}^c)} \sum_{n=1}^{\mathcal{C} (\mathcal{N}_{\mathcal{T}^t}^c)} \mathcal{C} (\Phi(\mathcal{X}^n_\mathcal{T^\textit{tc}})) ||_2
\end{aligned}
\end{equation}
where $C$ is the number of categories, $\mathcal{Y_{S^\textit{c}}}\vert \mathcal{X_{S^\textit{c}}}$ (or $\mathcal{C} (\mathcal{Y}_{\mathcal{T}_{\mathcal{P}}^{tc}})\vert  \mathcal{C}(\mathcal{X}_{\mathcal{T}^{tc}}))$)  represents $c^{th}$ category  data in the source  domain or confident pseudo labeled target domain. $\mathcal{N}_\mathcal{S}^c$ or $\mathcal{C} (\mathcal{N}_{\mathcal{T}^t}^c$) is the number of samples in the  $c^{th}$ category in the source domain or confident pseudo labeled target domain in each $t$.

In each iteration, we first use the initial classifier to generate pseudo labels for the target domain. Then, the pseudo labels will be progressively refined. We empirically demonstrate that such iterative learning is effective and efficient in improving target domain accuracy.

\subsection{PRPL model}
The framework of our proposed PRPL model is depicted in Fig.~\ref{fig:model}.  Taken altogether, our model minimizes the following objective function:
\begin{gather}\label{eq:all}
\scalebox{0.95}{$
\begin{aligned}
& \mathcal{L}  (\mathcal{X_S}, \mathcal{Y_S},  \mathcal{X_T}) =  \mathop{\argmin}    (\mathcal{L_S}   +  \mathcal{L_{MMD}}  + \sum_{t=1}^{T} \mathcal{L}_{\mathcal{U}}^{t})
\end{aligned}$}
\end{gather}
%
where $\mathcal{L_S}$ is the source classification loss and  $\mathcal{L_{MMD}}$ minimizes the distance between initial source and target represented data. $\mathcal{L}_{\mathcal{U}}^{t}$ is the loss function of each $t$. $T$ represents the number of iterations of training.
\subsection{Domain Adaptation Theory}\label{sec:the}

We formalize the theoretical error bound of the target domain for proposed recurrent learning in Lemma 1.

\textbf{Lemma 1.} \textit{Let $h$ be a hypothesis in a hypothesis space $H$. $\epsilon_{\mathcal{S}} (h)$ and $\epsilon_{\mathcal{T}} (h)$ represent the source and target domain risk, respectively~\cite{ben2007analysis}.  We have
\begin{equation*}
\begin{aligned}\label{eq:ana}
\epsilon_{\mathcal{T}} (h)  & 
\leq \epsilon_{\mathcal{S}} (h) + d_{\mathcal{H}}( P(\Phi(\mathcal{X_S})), P(\Phi(\mathcal{X_T}) )) + \gamma, 
\end{aligned}
\end{equation*}
where $ d_{\mathcal{H}}( P(\Phi(\mathcal{X_S})), P(\Phi(\mathcal{X_T}) ))$ is the $\mathcal{H}$ divergence between the probability distribution of the source and target domain. $\gamma= \epsilon_{\mathcal{S}} (h^{*}, \mathcal{Y_S}) + \epsilon_{\mathcal{T}} (h^{*}, \mathcal{F}(\Phi(\mathcal{X_T})))$ is the adaptability to quantify the error in ideal hypothesis $h^{*}$ space of source and target domain, which should be small.}

During the recurrent pseudo-labeling, we expect the $\mathcal{H}$ divergence between the distributions of latent feature space can be minimized, and that an ideal hypothesis exists with low risk on both domains, which is corresponding to a small $\beta$ in Lemma 1. In addition,  such a divergence is assessed by $d_{\mathcal{H}}( P(\Phi(\mathcal{X_S})), P(\Phi(\mathcal{X_T}) ))$ $\approx  Dist^{Ma} +\frac{1}{T} \sum_{t=1}^{T}Dist^{Co}_t$. Therefore, with the implicitly minimized training risk, domain divergence, and the adaptability of true hypothesis $h$, the generalization bound of $\epsilon_{\mathcal{T}} (h) $ can be achieved.

\section{Experiments}
\subsection{Setup}
 \paragraph{Datasets.}\textbf{Office + Caltech-10} \cite{gong2012geodesic} consists of Office~10 and Caltech~10 datasets with 2,533 images from ten classes in four domains: Amazon (A), Webcam (W), DSLR (D) and Caltech (C). There are twelve tasks in Office + Caltech-10 dataset. \textbf{Office-31} \cite{saenko2010adapting} consists of 4,110 images in 31 classes from three  domains: Amazon (A), Webcam (W), and DSLR (D). We evaluate methods across all six transfer tasks. \textbf{Office-Home} \cite{venkateswara2017deep} contains 15,588 images from 65 categories. It has four domains: Art (Ar), Clipart (Cl), Product (Pr) and Real-World (Rw). There are also twelve tasks in this dataset.  Therefore, we have 30 tasks in our experiment.   In experiments, C$\shortrightarrow$A represents learning knowledge from domain C which is applied to domain A.

 \begin{table}[t]
\small
\begin{center}
\captionsetup{font=small}
\caption{Pre-trained feature mean distance (1.0e+05) between two domains and feature extraction time (minutes) for three datasets (MD: mean distance; OC10: Office + Caltech-10; IR: Inceptionresnetv2; EB7: EfficientNetB7; NM: Nasnetmobile). }
\vspace{-0.3cm}
\setlength{\tabcolsep}{+1.3mm}{
\begin{tabular}{lll|ll|lll}
\hline \label{tab:md}
 \multirow{2}{*}{Networks}
 &  \multicolumn{2}{c}{OC10} & \multicolumn{2}{c}{Office-31} & \multicolumn{2}{c}{Office-Home}\\
 \cmidrule{2-7}
& MD & Time & MD & Time & MD & Time \\
\hline
SqueezeNet~\cite{iandola2016squeezenet} & 41.93& 0.79
& 32.61 & 1.16 &28.38 &7.23 \\
AlexNet~\cite{krizhevsky2012imagenet} & 20.05& 0.29 &   18.74&0.46   &  19.68 &4.20 \\
GoogleNet~\cite{szegedy2015going}  & 15.25&0.28 &    15.74&0.47   &  13.89 &4.25 \\
ShuffleNet~\cite{zhang2018shufflenet}  &24.97&0.32   &  25.81&0.54  & 21.82 &4.47  \\
ResNet18~\cite{he2016deep}   & 19.27&0.29   &  18.99&0.48  &   17.27  & 4.30 \\
Vgg16~\cite{simonyan2014very}  & 15.92 &0.40  &   16.11&0.64   &  16.70 &4.80  \\
Vgg19~\cite{simonyan2014very}  &15.49&0.43  &   16.17&0.68   &  16.86 &4.94 \\
MobileNetv2~\cite{sandler2018mobilenetv2} & 8.53&0.34  &   8.13&0.57   &  8.17 &4.62 \\
NM~\cite{zoph2018learning} &6.03&0.90   &  5.44&1.21  &   6.66 & 6.21 \\
ResNet50~\cite{he2016deep}  & 18.94&0.39  &   19.62& 0.64  &   18.13  &4.78 \\
ResNet101~\cite{he2016deep} & 20.25&0.48   &  20.17&0.75  &  19.11 &5.18  \\
DenseNet201~\cite{huang2017densely} & 22.1&1.32   &  22.05&2.04 &    18.80 &9.56  \\
Inceptionv3~\cite{szegedy2016rethinking}  & 5.74&0.43  &   5.47&0.68  &   5.94&4.86 \\
Xception~\cite{chollet2017xception}  &6.02&0.70  &  5.75&1.13  &   7.03& 6.76\\
IR~\cite{szegedy2017inception}  & 5.40&0.81  &   5.73  & 1.19&  6.80 & 6.29 \\
NasnetLarge~\cite{zoph2018learning}  & 4.19&2.45  &4.04&3.64  &   6.15 & 14.65 \\
\hline
\textbf{EB7}~\cite{zoph2018learning}  & \textbf{1.13}& 4.06 &   \textbf{1.27} & 8.31 &  \textbf{1.49} &  23.64\\
\hline
\end{tabular}}
\end{center}
\vspace{-.6cm}
\end{table}

\paragraph{Implementation details.}
As in Zhang and Davison \cite{zhang2020impact},
we first extract features from the last fully connected layer from 17 different networks for the three datasets. 
Parameters in recurrent pseudo labeling are  $T=3$ and $p_T=[0.5, 0.8, 0.9]$. Learning rate ($\epsilon = 0.001$), batch size (64), and number of epochs (9) are determined by performance on the source domain.  We compare our results with 12 state-of-the-art methods. For fair comparison, we highlight those methods in bold that are re-implemented using our extracted features, and other methods are directly reported from their original papers. Specifically, we modify the architecture of DAN~\cite{long2015learning}, DANN~\cite{ghifary2014domain} and DCORAL~\cite{sun2016deep}, and replace the feature extractor with the best pre-trained features, while maintaining original loss functions\footnote{Source code is available at: \url{https://github.com/YoushanZhang/Transfer-Learning/tree/main/Code/Deep/PRPL}.}. Experiments are tested with a GeForce 1080 Ti.

\begin{table*}[t]
\small
\begin{center}
\caption{Accuracy (\%) on Office + Caltech-10 dataset}
\vspace{-0.3cm}
\setlength{\tabcolsep}{+1.6mm}{
\begin{tabular}{rccccccccccccc}
\hline \label{tab:OC+10}
Task & C$\shortrightarrow$A &  C$\shortrightarrow$W & C$\shortrightarrow$D & A$\shortrightarrow$C & A$\shortrightarrow$W & A$\shortrightarrow$D & W$\shortrightarrow$C & W$\shortrightarrow$A & W$\shortrightarrow$D & D$\shortrightarrow$C & D$\shortrightarrow$A & D$\shortrightarrow$W & \textbf{Ave.}\\
\hline
\textbf{DAN~\cite{long2015learning}}	&	\textbf{96.8} &	95.9 &	96.2	&	93.1 &	88.8 &	92.4 &	94.5 &	95.4 &	\textbf{100}	&89.1	&	95.9 &	95.9 &	94.5 \\
\textbf{DANN~\cite{ghifary2014domain}}	& 96.7	& 94.9 & 	97.5 & 	95.8 & 	94.9 & 	91.1 & 	94.7 & 	94.5 & 	\textbf{100}& 	94.9 & 	92.4 & 	93.9 & 	95.1 \\
\textbf{DCORAL~\cite{sun2016deep}}	&	96.5 &	97.6 &	96.8	&	96.3 &	98.3	& 96.8 &	94.9 &	95.8 &		\textbf{100}	& 94.6	&	95.8 &	99.0 &		96.9 \\
\hline
\hline
DDC~\cite{tzeng2014deep}	&	91.9&	85.4&	88.8&		85.0&	86.1&	89.0&	78.0&	83.8&		\textbf{100}	&	79.0	&	87.1&	97.7	&	86.1\\
RTN~\cite{long2016unsupervised} &93.7 & 96.9 &94.2 &88.1 &95.2 & 95.5& 86.6& 92.5& \textbf{100} & 84.6& 93.8 & 99.2 &93.4 \\
MDDA~\cite{rahman2020minimum} &93.6 & 95.2 &93.4 &89.1 &95.7 & 96.6& 86.5&94.8 & \textbf{100} & 84.7& 94.7 & 99.4 & 93.6\\
 \hline
  \hline
 PRPL$_{t=0}$& 96.7 & 98.6 &   98.1 &  95.5 & 99.0 & \textbf{100} & 95.3 &   \textbf{96.7} &  \textbf{100} &  95.1 &   \textbf{96.2} &   \textbf{99.7} &   97.6\\  
 PRPL$_{t=1}$& 96.7 & 98.6 &   99.4 &   96.2 & \textbf{99.3} & \textbf{100} & 96.4 &  96.6 &  \textbf{100} &  96.1&   96.1&  \textbf{99.7} &   97.9\\
 PRPL$_{t=2}$& \textbf{96.8} & 98.6 &   \textbf{100} &   96.4 & \textbf{99.3} & \textbf{100} & 96.4 &   96.5 &  \textbf{100} &  96.1 &   \textbf{96.2} &   \textbf{99.7} &   98.0\\
 \textbf{PRPL}$_{t=3}$& 96.7 & \textbf{99.0} &   \textbf{100} &   \textbf{96.6} & \textbf{99.3} & \textbf{100} & \textbf{96.6} &   96.6 &  \textbf{100} &  \textbf{96.2 }& \textbf{96.2}&   \textbf{99.7} &   \textbf{98.1}\\
  \hline
\end{tabular}}
\vspace{-.3cm}
\end{center}
\end{table*}

\begin{table*}[h!]
\begin{center}
\small  
\caption{Accuracy (\%) on Office-Home dataset}
\vspace{-0.3cm}
\setlength{\tabcolsep}{+0.9mm}{
\begin{tabular}{rccccccccccccc}
\hline \label{tab:OH}
Task & Ar$\shortrightarrow$Cl &  Ar$\shortrightarrow$Pr & Ar$\shortrightarrow$Rw & Cl$\shortrightarrow$Ar & Cl$\shortrightarrow$Pr & Cl$\shortrightarrow$Rw & Pr$\shortrightarrow$Ar & Pr$\shortrightarrow$Cl & Pr$\shortrightarrow$Rw & Rw$\shortrightarrow$Ar & Rw$\shortrightarrow$Cl & Rw$\shortrightarrow$Pr & \textbf{Ave.}\\
\hline
\textbf{DAN~\cite{long2015learning}} &	55.0 &72.5 &79.1 & 66.5 & 72.9 &74.3 &69.7 &57.4 &82.7 & 76.1 & 59.1 &85.4 & 70.9 \\
\textbf{DANN~\cite{ghifary2014domain}}	& 57.1	& 75.7 & 	80.3 & 	69.2 & 	76.7 & 	74.3 & 70.9 & 58.0 & 83.0 & 	77.8 & 59.8 & 	87.0 & 	72.5 \\
\textbf{DCORAL~\cite{sun2016deep}} & 59.1 &78.4 &82.3 & 71.1 & 79.0 &77.4 &71.2 &57.8 &84.3 & 78.7 & 60.4 &87.0 & 73.9 \\ 
\hline
\hline
CDAN-RM~\cite{long2018conditional}	& 49.2& 	64.8& 	72.9& 	53.8& 	62.4& 	62.9& 	49.8& 	48.8& 	71.5& 	65.8& 	56.4& 	79.2& 	61.5\\
CDAN-M~\cite{long2018conditional}	& 50.6& 	65.9& 	73.4& 	55.7& 	62.7& 	64.2& 	51.8& 	49.1& 	74.5& 	68.2& 	56.9& 	80.7& 	62.8\\
ETD~\cite{li2020enhanced} &51.3 & 71.9& 85.7& 57.6 &69.2 &73.7 &57.8 &51.2 &79.3 &70.2 &57.5 &82.1 &67.3 \\
TADA~\cite{wang2019transferable} & 53.1 &72.3& 77.2& 59.1 &71.2& 72.1& 59.7& 53.1& 78.4 &72.4& 60.0 &82.9& 67.6 \\
SymNets~\cite{zhang2019domain}& 47.7 & 72.9 & 78.5 & 64.2  & 71.3  &74.2  & 64.2  & 48.8 &  79.5&  74.5 &52.6 & 82.7& 67.6 \\
\hline
\hline
 PRPL$_{t=0}$& 65.0 & 83.2 &   88.3 &   78.0 & 83.9 & 85.2  &   75.3 & 65.3 &  87.6 &   82.1 &   66.2 &   90.6 &79.2\\
 PRPL$_{t=1}$& 67.0 & 84.3 &   \textbf{89.4} &   79.4 & 85.2 & 85.7  &   78.1 &  68.1 &  88.5 &  83.4 &   68.6 &  91.3 &80.8\\
 PRPL$_{t=2}$& \textbf{68.0} & \textbf{84.5} &   \textbf{89.4} &   79.5 & \textbf{85.7 }& 86.0  &   79.1 &  68.9 &  88.6 &   83.3 &   68.5 &   91.4 & 81.1\\
 \textbf{PRPL}$_{t=3}$& 67.6 & \textbf{84.5} &   \textbf{89.4} &   \textbf{79.8} & \textbf{85.7} & \textbf{86.3} &  \textbf{79.2} &  \textbf{69.1} &  \textbf{88.7} &   \textbf{83.8} &  \textbf{68.9} &   \textbf{91.5} &\textbf{81.2}\\
\hline
\end{tabular}}
\end{center}
\vspace{-0.8cm}
\end{table*}


\begin{table}[!htb]
\small
      \caption{Accuracy (\%) on Office-31 dataset}
      \vspace{-0.3cm}
      \centering
\setlength{\tabcolsep}{+0.1mm}{
\begin{tabular}{rcccccccc|c|c|c|c|c|c|c|c|}
\hline \label{tab:O31}
Task & A$\shortrightarrow$W &  A$\shortrightarrow$D & W$\shortrightarrow$A & W$\shortrightarrow$D & D$\shortrightarrow$A & D$\shortrightarrow$W  & \textbf{Ave.}\\
\hline
\textbf{DAN~\cite{long2015learning}} &	85.8 &	88.0 &	75.7	&98.4 &	74.7 &	95.0 &	86.3 \\
\textbf{DANN~\cite{ghifary2014domain}} &	85.9 &	90.2 &	76.6 &	98.8 &	78.2 &	96.2 & 87.7\\
\textbf{DCORAL~\cite{sun2016deep}} &90.7 &	90.6 &	79.0 &	98.8 &	78.7 &	97.1 & 89.2\\
\hline
\hline
RTN~\cite{long2016unsupervised} &	84.5 &	77.5 &	64.8 &	99.4 &	66.2 &	96.8 &	81.6 \\
ETD~\cite{li2020enhanced} &92.1 & 88.0 & 67.8 & \textbf{100} & 71.0 & \textbf{100} &  86.2 \\
TADA~\cite{wang2019transferable} &94.3 & 91.6  & 73.0  &99.8  & 72.9 & 98.7 & 88.4 \\
SymNets~\cite{zhang2019domain}& 90.8& 93.9  &72.5  & \textbf{100} & 74.6 & 98.8& 88.4 \\
CAN~\cite{kang2019contrastive} & 94.5 & 95.0  &77.0   &99.8  & 78.0 & 99.1 &  90.6 \\
\hline
\hline
PRPL$_{t=0}$ & 92.1 &	96.0 &	80.4 &	98.6 &	80.1 &	96.1 &	90.6	\\
PRPL$_{t=1}$ & 95.4 &	\textbf{97.0 }&	81.8 &	99.2&	82.1 &	97.0 &	92.1	\\
PRPL$_{t=2}$ & 95.6 &	96.8 &	82.2 &	99.2&	82.8 &	97.1 &	92.3	\\
\textbf{PRPL$_{t=3}$} & \textbf{95.9} &	\textbf{97.0} &	\textbf{82.4} &	99.2&	\textbf{83.0} &	97.1 &\textbf{92.4}	\\
\hline
\end{tabular}}
\vspace{-0.3cm}
\end{table}

\begin{table}[h]
\begin{center}
\caption{Ablation experiments on Office-31 }
\vspace{-0.3cm}
\setlength{\tabcolsep}{+0.5mm}{
\begin{tabular}{llllllll}
\hline \label{tab:ab}
Task & A$\shortrightarrow$W &  A$\shortrightarrow$D & W$\shortrightarrow$A & W$\shortrightarrow$D & D$\shortrightarrow$A & D$\shortrightarrow$W  & \textbf{Ave.}\\
\hline
PRPL$_{t=0}$-$\mathcal{M}$	& 91.6 &	96.0 &	79.9 &	98.8 &	80.6 &	96.0 &	90.5\\
PRPL$_{t=1}$-$\mathcal{M}$	& 94.1 &	96.6 &	81.4 &	\textbf{99.2} &	82.5 &	97.0 &	91.8\\
PRPL$_{t=2}$-$\mathcal{M}$	& 94.2 &	96.3 &	81.5 &	\textbf{99.2} &	82.9 &	97.0 &	91.9\\
PRPL$_{t=3}$-$\mathcal{M}$ & 94.3 & 96.4 & 81.9  & \textbf{99.2} & 82.9 & 97.0	& 92.0	\\
\textbf{PRPL$_{t=3}$} & \textbf{95.9} &	\textbf{97.0} &	\textbf{82.4} &	\textbf{99.2} &	\textbf{83.0} &	\textbf{97.1} &	\textbf{92.4}	\\
\hline
\end{tabular}}
\end{center}
\vspace{-0.6cm}
\end{table}

\subsection{Results}
\paragraph{Pre-trained feature selection.}\label{sec:pre}
We first conduct experiments to select the best pre-trained features for both the source and target domains. We calculate the distance between two domains using the aforementioned pre-trained feature distance in Eq.~\ref{eq:pre}. The smallest distance between two domains reveals the corresponding pre-trained network that is the best for feature extraction.  In Tab.~\ref{tab:md}, we first report the mean distance between two domains of three datasets (for Office + Caltech-10 and  Office-Home, the mean distance is the average of twelve tasks. For Office-31, the mean distance is the average of six tasks). It is obvious that pre-trained features from EfficientNetB7 have the smallest distance between two domains, which suggests the EfficientNetB7 is the best deep network compared to the other 16 networks. In addition, this observation is consistent with~\cite{zhang2020impact}, that a better ImageNet model will produce better pre-trained features for UDA. Furthermore, we also list the computation time for feature extraction of each dataset. We notice that EfficientNetB7 consumes more time than other networks since it has more complex layers, and it needs more memory to process images. However, the longest time is 23.64 minutes, and these features will not be extracted again during the training. Therefore, we opt for  EfficientNetB7 as the feature extractor to extract pre-trained features for the benchmark datasets. We then use the extracted features $\Phi(\mathcal{X_S})$ and $\Phi(\mathcal{X_T})$ to perform the domain transfer tasks. In \textbf{supplementary material}, we also validate the effectiveness of the proposed distance function $Dist_k^{Pre}$ in choosing the best pre-trained features by comparing to MMD and mean cosine distance function.

\begin{figure*}[t]
\centering
\begin{subfigure}{0.49\textwidth}
\includegraphics[width=\linewidth]{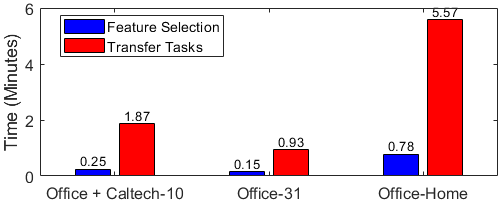}
\caption{Time of three datasets  } \label{fig:time_all}
\end{subfigure}
\begin{subfigure}{0.49\textwidth}
\includegraphics[width=\linewidth]{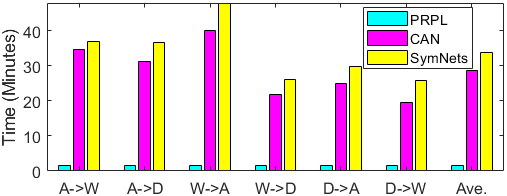}
\caption{Time of each task in Office31} \label{fig:time_o31}
\end{subfigure} 
\vspace{-0.3cm}
\caption{Computation time comparison. (a) is the total computation time, that includes the pre-trained feature selection and all transfer tasks (twelve for Office + Caltech-10, six for Office-31 and another twelve for Office-Home). (b) compares the PRPL model with the other two baselines in each task of Office31 (feature extraction time is also included). On average, our PRPL model is approximately 18 times faster than CAN~\cite{kang2019contrastive}, and 21 times faster than SymNets~\cite{zhang2019domain}.  } \label{fig:time}
\vspace{-0.4cm} 
\end{figure*}
\vspace{-.3cm}
\paragraph{Domain transfer accuracy.}
The performance on Office + Caltech-10, Office-Home and Office-31 are shown in Tables~\ref{tab:OC+10}-\ref{tab:O31}.  Our PRPL model outperforms all state-of-the-art methods in terms of average accuracy (especially in the Office-Home dataset). It is compelling that our PRPL model substantially enhances the classification accuracy on difficult adaptation tasks (e.g., W$\shortrightarrow$A task in the Office-31 dataset and the challenging Office-Home dataset, which has a larger number of categories and different domains are visually dissimilar).  Our model also outperforms three re-implemented baselines (DAN, DANN, and DCORAL), which use the same EfficientNetB7 features as our model. 
   
In Office + Caltech-10, although the final accuracy in recurrent learning is 98.1\%, it does not improve much from 97.6\% to 98.1\% (from first recurrent learning to the third recurrent learning). One reason is the classification accuracy is high (more than 97\%). It is hence difficult to make  a  large improvement).  However,  our model still provides a 1.2\%  improvement over the best baseline (DCORAL). The mean accuracy on the Office-31 dataset is increased from 90.6\% to 92.4\%. We notice that accuracy is obviously improved when $t=1$, and then refined when $t=2/3$.  A similar tendency is observed on  Office-Home dataset. Therefore, the best pre-trained features are powerful, and the recurrent learning is effective in improving the classification accuracy. In addition, we also show the computation time of our proposed PRPL model in Fig.~\ref{fig:time}. We require relatively less computation time of all three datasets in Fig.~\ref{fig:time_all}. In particular, we compare the computation time of our model with two other methods (CAN~\cite{kang2019contrastive}  and SymNets~\cite{zhang2019domain}) on each task in Office-31 datasets. Our model is obviously faster than the other two (18 and 21 times faster). Therefore, our PRPL model is fast and accurate 
in UDA tasks. 

\subsection{Ablation study}
To demonstrate the effects of  $\mathcal{L_{MMD}}$  loss  on classification  accuracy, we present an ablation study in Tab.~\ref{tab:ab}. We observe that $\mathcal{L_{MMD}}$ loss is useful in improving performance during each training iteration comparing with Tab.~\ref{tab:O31}. 

\subsection{Parameter Analysis}
\begin{figure*}[t]
\centering
\begin{subfigure}{0.49\textwidth}
\includegraphics[width=\linewidth]{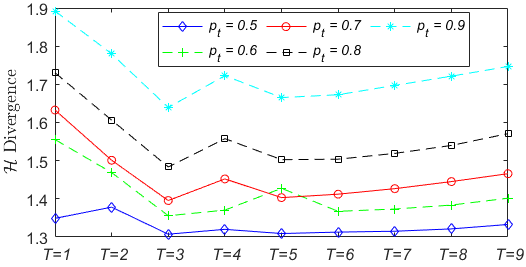}
\caption{Effect of different $T$ on $d_{\mathcal{H}}$ } \label{fig:ima}
\end{subfigure}
\begin{subfigure}{0.49\textwidth}
\includegraphics[width=\linewidth]{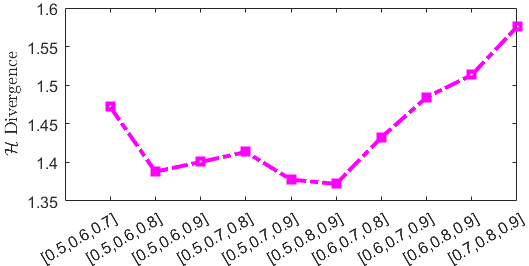}
\caption{Effect of different $p_t$ on $d_{\mathcal{H}}$} \label{fig:imb}
\end{subfigure} 
\vspace{-0.3cm}
\caption{Parameter analysis for $T$ and $p_t$. In (a), $d_{\mathcal{H}}$ is minimum when $T=3$. In (b), the x-axis denotes different $p_t$, and in each array, it contains $p_1, p_2$ and $p_3$ since $T=3$. $d_{\mathcal{H}}$ is minimum when $p_T=[0.5, 0.8, 0.9]$.  } \label{fig:rela}
\vspace{+0.1cm}
\end{figure*}
There are two hyperparameters $T$ and $p_t$ in PRPL that control the number of recurrent learning repetitions and the probability of selecting the confident examples. To get the optimal parameters, we randomly opt the task Rw$\shortrightarrow$Pr and run a set of experiments regarding different values of each parameter. Notice that it is inappropriate to tune parameters using the target domain accuracy since we do not have any labels in the target domain. Therefore, we report the $\mathcal{H}$ divergence between two domains, as stated in Sec.~\ref{sec:the}.  Since $\mathcal{H}$ divergence can be assessed by $d_{\mathcal{H}}( P(\Phi(\mathcal{X_S})), P(\Phi(\mathcal{X_T}) )) \approx  Dist^{Ma} +\frac{1}{T} \sum_{t=1}^{T}Dist^{Co}_t$, we calculate the marginal distance and recurrent conditional distance to tune these two parameters. $T$ is selected from $\{1,2,3,4,5,6,7,8,9\}$, and $p_t$ is selected from $\{0.5,0.6,0.7,0.8,0.9\}$, we fix one parameter and vary another one at a time. 

Results presented in Fig.~\ref{fig:rela} demonstrate that our PRPL model is not very sensitive to a wide range of parameter values since the $\mathcal{H}$ divergence ($d_{\mathcal{H}}$) is not significantly changed. In Fig.~\ref{fig:ima}, we first tune the parameter $T$ across five different $p_t$, and it consistently shows that $d_{\mathcal{H}}$ achieves the minimum  value when $T = 3$. Therefore, the hyperparameter $T = 3$ is the best since the discrepancy between two domains is minimized. After fixing $T$, we then present the effect of different $p_t$ on $d_{\mathcal{H}}$ in Fig.~\ref{fig:imb}. When $p_1 =0.5, p_2=0.8$ and $p_3=0.9$, $d_{\mathcal{H}}$ achieves the minimum  value. 
In Fig.~\ref{fig:rela}, a large $p_t$ tends to have a larger $d_{\mathcal{H}}$ (e.g., $p_t=0.9$ in Fig.~\ref{fig:ima} and $p_T=[0.7, 0.8, 0.9]$ in Fig.~\ref{fig:imb}) since a larger $p_t$ will select relatively fewer examples during the training \textit{i.e.,} $\mathcal{N_U}$ is small, and the discrepancy between two domains cannot be well minimized. Therefore, the parameter analysis is useful in finding the best hyperparameters for our PRPL model.

\begin{figure*}[h!]
\centering
\includegraphics[width=2\columnwidth]{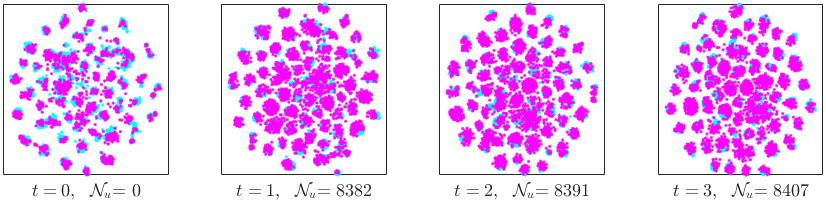}
\caption{t-SNE of feature visualization of the task Rw$\shortrightarrow$Pr in our recurrent pseudo-labeling learning when $T=3$. Our PRPL model improves the consistency of representations across domains. Also, the number of updated label domain $\mathcal{N_U}$ is growing with the increasing of time $t$ (magenta color: source domain, cyan color: target domain). Best viewed in color.}
\label{fig:tsne}
\end{figure*}

\subsection{Feature Visualization}
To further investigate the quality of invariant representation learned during the transition from the source domain to the target domain, Fig.~\ref{fig:tsne} visualizes embeddings of the task Rw$\shortrightarrow$Pr in the Office-Home dataset. In this figure, the magenta dots represent the source domain, and the cyan dots denote the target domain. We can observe that the representation becomes more discriminative when $t=1$, compared with $t=0$ (no recurrent learning). Although the representations of $t=2$ and $t=3$ are slightly improved, PRPL keeps producing confident examples 
in $\mathcal{D}_\mathcal{U}^t$ since $\mathcal{N}_\mathcal{U}^t$  is increasing.

\section{Discussion}
\paragraph{What can we learn from PRPL?}

Recurrent learning is effective and accurate to improve target domain accuracy. The architecture of our proposed PRPL is neat and straightforward. However, our model outperforms state-of-the-art methods and achieves the highest accuracy so far. There are two compelling advantages: 1) we extract pre-trained features from 17 well-trained ImageNet networks, and we select the best pre-trained features based on the domain distance.  EfficientNetB7 produces high-quality features for the datasets. 2) the proposed recurrent pseudo labeling effectively keeps improving the target domain accuracy in each iteration. Therefore, the generated confident pseudo labels are useful in updating the network parameters, which further reduces the domain discrepancy. 
\vspace{-0.3cm}
\paragraph{Is a pre-trained ImageNet model helpful?}

Yes. ImageNet pre-training is important in improving the quality of extracted features. Most existing work focused on fine-tuning the ResNet50 network to perform domain transfer tasks. One underlying reason is that ResNet50 is not a very complex model, and it is easier to re-train the network without the need of multiple GPUs. However, as we can see from the results, the pre-trained features are efficient and easy to use. Therefore, we recommend selecting better pre-trained features for UDA.

\section{Conclusion}
In this paper, we  efficiently and effectively select the best pre-trained features among seventeen well-trained ImageNet models in an unsupervised fashion. The EfficientNetB7 model shows the highest quality in extracting image features. We then propose recurrent pseudo-labeling training to progressively generate confident labels for the target domain. Extensive experiments demonstrate that the proposed PRPL model achieves superior accuracy, noticeably higher than state-of-the-art domain adaptation methods.

{\small
\bibliographystyle{ieee_fullname}
\bibliography{egbib}
}

\end{document}